\documentclass[review]{elsarticle}

\usepackage{hyperref,amsmath,graphicx,multirow,array,bm,booktabs,amssymb,bookmark,algorithm,algorithmic}

\journal{Signal Processing: Image Communication}
\begin{document}

\begin{frontmatter}

\title{Kernel Based Low-Rank Sparse Model for Single Image Super-Resolution}
\tnotetext[mytitlenote]{This research is supported by the National Natural Science Foundation
of China grant numbers [61572395, 61601362, 61133008].}

\author{Jiahe Shi\corref{mycorrespondingauthor}}
\ead{sjh19900831@163.com}
\author{Chun Qi\corref{mycorrespondingauthor}}
\ead{qichun@mail.xjtu.edu.cn}

\address{School of Electronic and Information Engineering,Xi'an Jiaotong University\corref{mycorrespondingauthor}}

\cortext[mycorrespondingauthor]{Corresponding author}

\address{Xi'an, Shaanxi 710049 China}

\begin{abstract}
Self-similarity learning has been recognized as a promising method for single image super-resolution (SR) to produce high-resolution (HR) image in recent years. The performance of learning based SR reconstruction, however, highly depends on learned representation coefficients. Due to the degradation of input image, conventional sparse coding is prone to produce unfaithful representation coefficients. To this end, we propose a novel kernel based low-rank sparse model with self-similarity learning for single image SR which incorporates nonlocal-similarity prior to enforce similar patches having similar representation weights. We perform a gradual magnification scheme, using self-examples extracted from the degraded input image and up-scaled versions. To exploit nonlocal-similarity, we concatenate the vectorized input patch and its nonlocal neighbors at different locations into a data matrix which consists of similar components. Then we map the nonlocal data matrix into a high-dimensional feature space by kernel method to capture their nonlinear structures. Under the assumption that the sparse coefficients for the nonlocal data in the kernel space should be low-rank, we impose low-rank constraint on sparse coding to share similarities among representation coefficients and remove outliers in order that stable weights for SR reconstruction can be obtained. Experimental results demonstrate the advantage of our proposed method in both visual quality and reconstruction error.
\end{abstract}
\end{frontmatter}

\section{Introduction}

\par High resolution (HR) images are generally preferred to low resolution (LR) ones in many applications of computer vision, such as remote sensing, medical imaging and video surveillance. However, the resolution is always limited by the constraint of optical imaging systems and hardware devices. As a software technique to break this limitation, super-resolution (SR) has been developed to reconstruct HR images from the observed LR ones using specific algorithms. SR methods can be divided into two categories: reconstruction based and learning based methods.

\par Reconstruction based methods recover HR images with help of prior knowledge and statistics of natural images, such as gradient profile prior \cite{Sun}, Gaussian mixture model \cite{GMM}, wavelet based model \cite{wavelet} and total variation (TV) \cite{TV}. Global constraint \cite{Irani} has also been widely used as a typical back-projection technique for SR.

\par Learning based methods predict the missing HR details by learning the model of relationships between pairs of LR and HR examples. The representative methods include neighbor embedding (NE) algorithm \cite{Chang}, sparse coding (SC) based method \cite{Yang2010} and position-based method \cite{Ma}. As for the above mentioned learning based methods, external examples from training images are required.
 \par However, internal examples instead of external ones can also be utilized. In nature images, sufficient examples which are highly correlated to the input patches can be found in the input image, its repeatedly down-sampled and subsequently up-scaled versions. In the past few years, self-similarity has been successfully utilized for SR\cite{Glasner,ZhangK2013,Yu2014,Bevilacqua2014}. Glasner et al. \cite{Glasner} first designed an appealing self-similarity learning framework. With the help of self-examples, the input image is repeatedly magnified to the desired size. By this coarse-to-fine strategy, the difficulty of each step is alleviated, which benefits the performance of the whole SR system. Due to these advantages, self-similarity learning has been followed by many researchers in recent years.  Bevilacqua et. al \cite{Bevilacqua2014} proposed a new double pyramid SR model with simple multi-variate regression to learn the direct mappings between LR and HR patches. Zhang et. al \cite{ZhangK2013} presented a neighbor embedding based self-similarity learning SR scheme with spatially nonlocal regularization. Yu et. al \cite{Yu2014} combined the self-similarity learning with sparse representation to perform SR. However, due to the degradation (i.e. blurring and down-sampling) of the observed image which is also the source of self-examples. The above mentioned conventional learning methods are prone to produce unfaithful representation coefficients, which are not suitable for accurate SR reconstruction. To solve the problem, a nonlocally constrained learning methods have been introduced recently \cite{Dong2013}. Dong et. al \cite{Dong2013} proposed a nonlocally centralized sparse representation for image restoration using PCA dictionary trained from self-examples. Specifically, they defined the deviation of the learned sparse codes from the expected true ones as sparse coding noise (SCN). As suggested by their work, SR performance can be improved by suppressing SCN through calculating nonlocal means of the sparse codes for similar neighbors of the LR input patches as an estimation of the optimal codes. However, the estimation is still a weighted linear combination of codes for the similar patches. In our previous work\cite{prework}, we found that nonlinear low-rank constraint can be used to suppress SCN in self-similarity learning scheme for SR. Furthermore, in this paper, we propose a novel kernel based low-rank sparse coding (KLRSC) method via self-similarity learning for single image SR. Self-examples are extracted from the input image itself, its degraded versions and up-scaled ones. The input image is gradually super-resolved. In each magnification, similar column components of a nonlocal data matrix which consists of a vectorized input patch and its nonlocal neighbors can be observed. This property of similarity leads to the nature of low-rank. Furthermore, we also use the kernel method \cite{kernel1997} to capture nonlinear structures of data, the nonlocal data are mapped into a high-dimensional feature space by kernel method. In our work, we find that the low-rank property is preserved when the nonlocal patches are mapped into the kernel space. Due to this observation, we assume that the sparse codes for nonlocal matrices should be approximately low-rank. Thus, we perform kernel based low-rank sparse coding to gain accurate coefficients for self-similarity learning based SR. Experimental results demonstrate the advantage of our proposed method in both visual quality and reconstruction error. Our contributions are two folds:
\par
\begin{enumerate}
\item The low-rank property is proved to be preserved when the nonlocal data are mapped into high dimensional space by kernel method.
\item A novel kernel based low-rank sparse coding based scheme for single image SR is proposed, which exploits both low-rank property and nonlinear structural information of nonlocal data in a high-dimensional space.
\end{enumerate}
\par The remainder of this paper is organized as follows. In Section 2, we describe the proposed method in detail. The experimental results are given in Section 3. We conclude this paper in Section 4.
\par Our preliminary work has appeared in \cite{prework}.

\section{Proposed kernel based low-rank sparse model for single image SR}
\subsection{Overview}
\par In this section, we start the discussion of the kernel based low-rank sparse model for single image SR. We adopt the double pyramid self-similarity learning framework which is the same as that of \cite{Bevilacqua2014}. By coarse-to-fine strategy, the observed image is zoomed in by several times to reach the expected size. In each magnification, we perform KLRSC to learn the representation coefficients for SR reconstruction. Both the self-examples, the structural information and the underlying nonlinear structure of nonlocal similar examples are exploited in the coding stage. Then, an interim image can be recovered by the learned coefficients and self-examples for the next magnification. When the image reaches the desired size, the iterations of magnification will terminate.

\subsection{Self-similarity learning and gradual magnification}
\par
In the stage of self-similarity learning, the pairs of examples are extracted from two pyramids of images. The flowchart of the double pyramids model is shown in Fig. \ref{fig:FlowChart}. We denote the pyramid which composes of the sequences of the input image and its several down-sampled versions as $I_{-n}$ for $n = 1,2,...N_D$. Given the input image $I_0$, the down-sampling is repeated for $N_D$ times with a factor of $s=1.25$ at each time. The $n$-th layer $I_{-n}$ is represented as:

\begin{figure}[ht!]
\centering
\setlength{\abovecaptionskip}{-0.5cm}
\includegraphics[width=\textwidth]{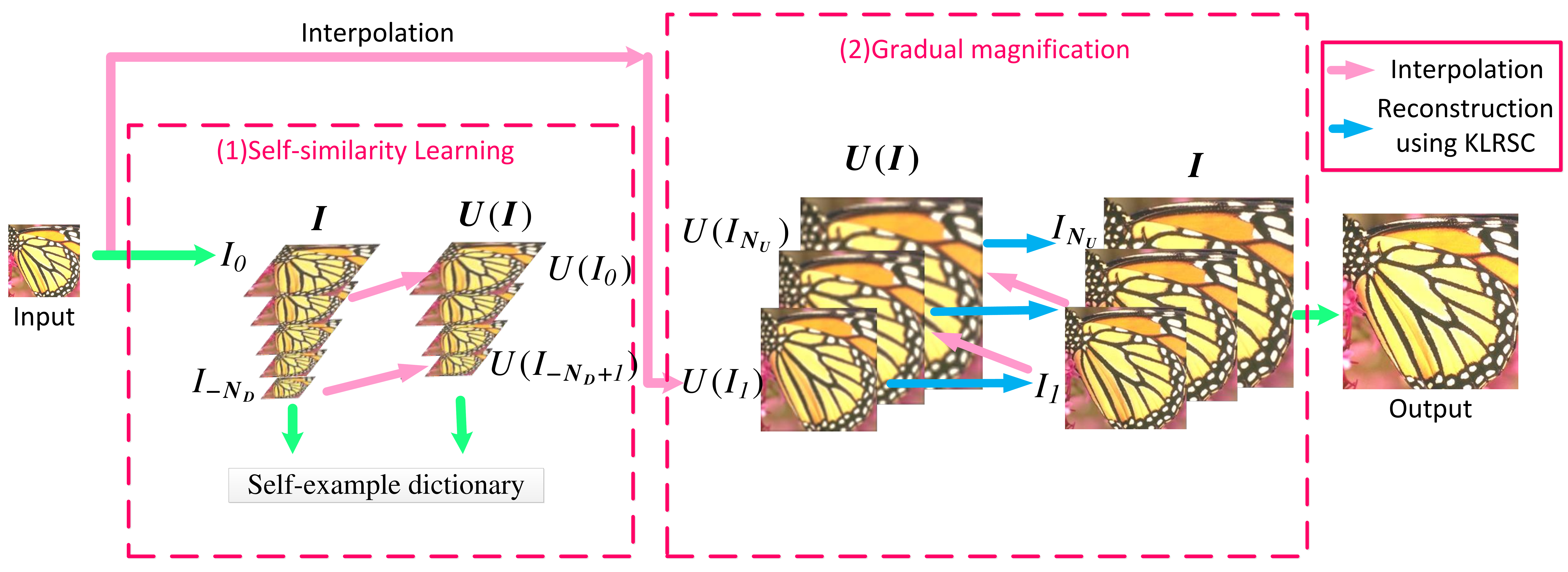}
\caption{ Overview of double pyramids model for single image SR. Sub-figure (1) with dashed lines stands for collection of self-examples . Sub-figure (2) with dashed lines denotes the gradual magnification with KLRSC for SR reconstruction.}
\label{fig:FlowChart}
\end{figure}

\begin{equation}
\label{ref:HR_pyramid}
I_{- n} = (I_0 * B_n)_{ \downarrow {s^n}}
\end{equation}
where $*$ is a convolution operator and $\downarrow {s^n}$ denotes the down-sampling operator by a factor of $s^n$. $B_n$ is a Gaussian blur kernel with a standard variance $\sigma^2_n$ which can be computed as \cite{Blur}:
\begin{equation}
\label{ref:Blur_kernel_variance}
\sigma_n^2=n \sigma^2 log(s)/log(p)
 \end{equation}
 The up-scaled pyramid $U(I_{ - n})$ by bicubic interpolation is established as:
\begin{equation}
\label{ref:LR_pyramid}
U(I_{ - n}) = (I_{ - n - 1})_{\uparrow s}
\end{equation}
where $U(I_{-n})$ is the $n$-th layer of the up-scaled pyramid with respect to the layer $I_{-n}$ and $\uparrow s$ is an up-scaling operator by a factor of s. In order to obtain the pairs of self-examples, layer $I_{-n}$ and the corresponding layer $U(I_{-n})$ are divided into overlapping patches. For each patch from  $U(I_{-n})$, we use four high-pass filters to extract its gradient feature of the first- and second-order gradients in both vertical and horizontal directions:
\begin{equation}
\label{ref:HP_filters}
f_1=[1,-1],f_2={f_1}^T,\\
f_3=[-1,2,-1],f_4={f_3}^T.
\end{equation}
The four high-pass filtered features are concatenated into vector as a descriptor of the patch. As for the corresponding patch from $I_{-n}$, we extract its intensity feature by subtracting its mean value. We collect these two kinds of features from all layers and normalize them to unit $\ell 2$-norm to construct the dictionary. Let $H=\{x_r^d\}_{r =1}^K \in \mathbb{R}^{b \times K}$ represent the dictionary for reconstruction and $L=\{ y_r^d \}_{r = 1}^K \in {\mathbb{R}^{4b \times K}}$ denote the one for learning.
\par
In multi-step magnification, we gradually super-resolve the input image $I_0$. Given the total up-scale factor $p$, we repeat the magnification for $N_U = ceil(log_s p)$ times, where $ceil(x)$ returns the nearest integer larger than $x$. In the $i$-th magnification for $i = 1 ,2 , ... N_U$, we produce the interim layer by:
\begin{equation}
\label{ref:LR_upscale}
 U(I_i) = (I_{i-1})_{\uparrow s}
 \end{equation}
 Then we partition the $i$-th layer $U(I_i)$ is into overlapped patches and convert them into a set of normalized gradient features $Y^i = \{ y_j \} _i$. We recover the corresponding patches $X_i = \{ x_j \} _i$ by kernel based low-rank sparse representation and reconstruct the layer $I_i$ by weighted average operation on the overlapped region, which we describe in the following subsection.

\subsection{Kernel based low-rank sparse representation for SR reconstruction}
In this subsection, we present how to recover the super-resolved layers from the interpolated ones using our proposed KLRSC. We first describe the algorithm of KLRSC for SR reconstruction. Then we present the post-processing procedure by the incorporation of iterative back projection (IBP) \cite{Irani} and pixel-wise autoregressive (AR) model regularization \cite{Dong2011} to improve the quality of reconstructed layer.
\par
\begin{figure*}[!ht]
\setlength{\abovecaptionskip}{-1cm}
\includegraphics[width=\textwidth]{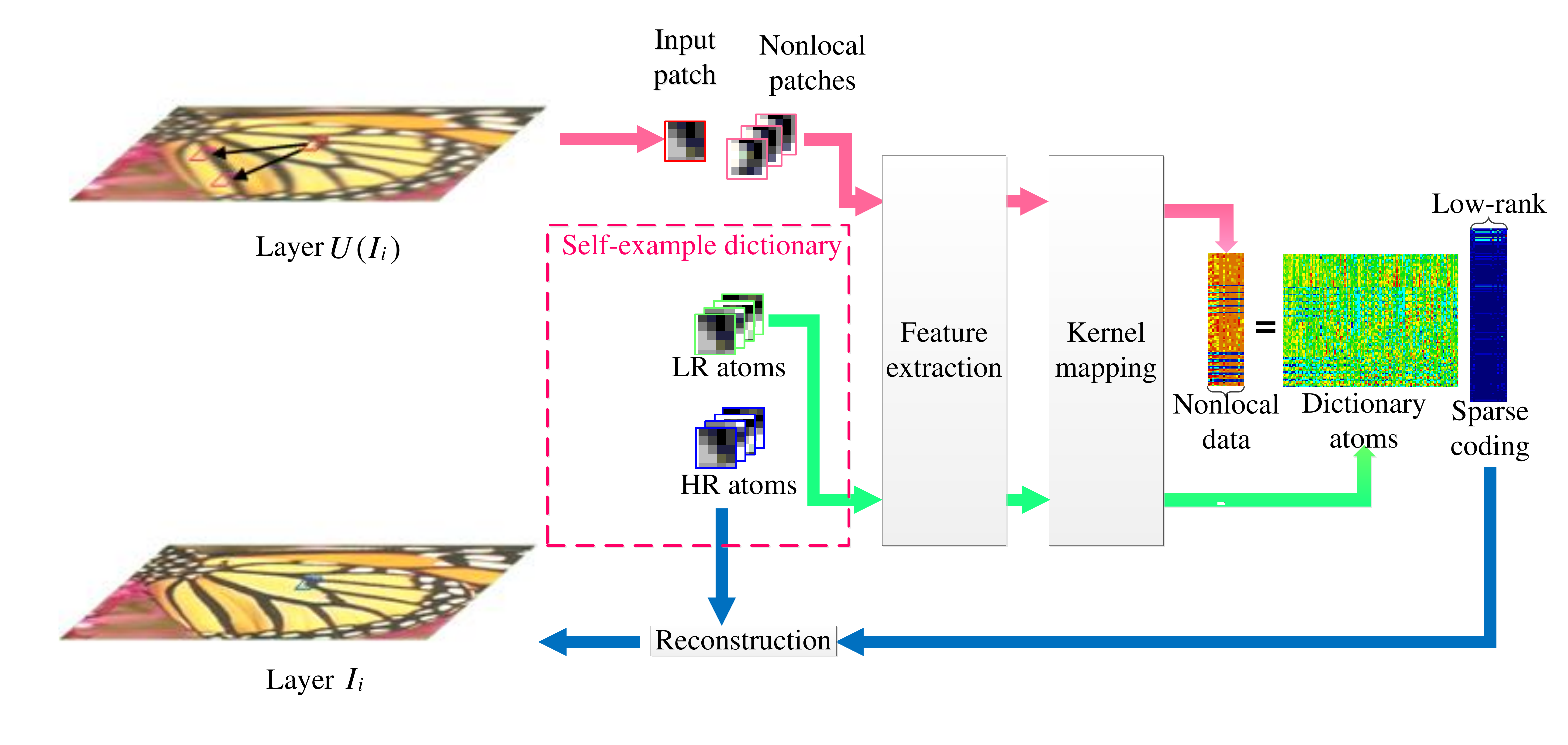}
\caption{KLRSC method for SR reconstruction.}
\label{fig:Lowrank}
\end{figure*}

Fig. \ref{fig:Lowrank} gives the illustration of KLRSC for SR reconstruction. Nonlocal-similarity is an effective prior for image reconstruction  \cite{ZhangK2013,Yu2014,Dong2013,Lu2014,NLSC,selfsimilarity2016}, which means that small patches tend to appear repeatedly at different locations of a natural image.  For each \emph{j}-th feature vector $y_j$ extracted from the \emph{i}-th layer $U(I_i)$, we select its ${K_N}$ most similar nonlocal neighbors in the same layer and stack them as columns $Y_j^N = \{ y_t \}_{t \in {G_N}(j)} \in {\mathbb{R}^{4b \times {K_N}}}$ where ${G_N}(j)$ refers to the indices of the nonlocal data. We also find its ${K_D}$ nearest atoms in the dictionary $L$ to create an subset $Y_j^D = \{ y_r^d \} _{r \in {G_D}(j)} \in {\mathbb{R}^{4b \times {K_D}}}$ for learning and accumulate the corresponding atoms in the dictionary $H$ to form the subset $X _j^D = \{ x_r\} _{r \in {G_D}(j)} \in {\mathbb{R}^{b \times {K_D}}}$ for reconstruction. ${G_d}(j)$ denotes the indices of the selected atoms.

\subsubsection{Implementation of KLRSC for SR}
In standard sparse coding, the sparse coding for the feature vector $y_j$ can be formulated as:
\begin{equation}
\label{ref:sparse coding}
{w } = {\rm{arg}}\mathop {{\rm{min}}}\limits_{{\alpha}} \frac{{\rm{1}}}{{\rm{2}}}{\rm{||}}{y_j - Y_j^Dw }{\rm{||}}_2^{\rm{2}}{+ \lambda ||}{w }{\rm{|}}{{\rm{|}}_{\rm{1}}}
\end{equation}
Different from the conventional approach, Recently, Zhang et. al \cite{Zhang2013} proposed a low-rank sparse coding method for image classification. They encoded densely sampled SIFT features in spatially local domain. The codes for spatially local features were assumed to be low-rank.
we introduce a low-rank constraint to regularize the representations for similar feature vectors. We attach the feature vector $y_j$ to the nonlocal feature vectors $Y_j^N$ to combine a grouped matrix $Y_j^A = [{y _j,Y_j^N}]$ with the nature of low-rank property. The corresponding sparse coefficient matrix for representing the data upon the subset $Y_j^D$ is also expected to be low-rank. Low-rank optimization relies on the proof that the convex envelope of rank is the nuclear norm under broad conditions \cite{lowrankproof}. Based on this theorem, low-rank optimization has been successfully used in many applications \cite{Candes2011,Liu2015,TILT,Tang2014,Ren2015,ICIPfusion} We use nuclear norm constraint \cite{lowrankproof} to formulate the low-rank optimization. The augmented optimization problem can be written as:
\begin{equation}
\label{ref:eqOrgProblem}
W_j = \mathop{\rm{argmin}}\limits_{W_j} \frac{{\rm{1}}}{{\rm{2}}}{\rm{||}}{Y_j^A} - {Y_j^D}{W_j}{\rm{||}}_{\rm{F}}^{\rm{2}} + {\rm{\lambda}}_{\rm{1}}  {\rm{||}}{W_j}{\rm{||}}_{\rm{1}} + {\rm{\lambda}}_{\rm{2}} {\rm{||}}{W_j}{\rm{||}}_{\rm{*}}
\end{equation}
where $W_j$ represents the corresponding weights that each atom in the subset $Y_j^D$ contributes in the reconstruction of the augmented data ${Y_j^A}$. The nuclear norm ${|| \cdot ||_*}$ is calculated by the sum of the matrix singular values, which is an approximation of rank. ${\lambda_1}$ and ${\lambda_2}$ are the parameters balancing different regularization terms.
\par
We also use the kernel method \cite{kernel1997} to capture the nonlinear structures of features, which can reduce the feature quantization error and improve the coding performance. As suggested by \cite{KSR,Shi2015}, we transform the augmented data ${Y_j^A}$ and the LR subset $Y_j^D$ into high dimensional space by the nonlinear mapping: $ \phi : R^{4b} \rightarrow R^F(4b<<F)$ to capture the relationship between them. The augmented features are transformed to $\phi(Y_j^A)$ and the corresponding LR subset is mapped to $\phi(Y_j^D)$. Given two column features $x$ and $y$, we define a kernel function $k(x,y)=\phi(x)^T\phi(y)$. In our work, we use Gaussian kernel function $k(x,y)=exp(-||x-y||_2^2/\sigma_G^2) (\sigma_G=1)$. Thus the kernel matrix $\phi(Y_j^D)^T\phi(Y_j^A)$ can be represented as $K_{Y_j^DY_j^A}$ where the element $(K_{Y_j^DY_j^A})_{m,n}=k(y_{j,m}^D,y_{j,n}^A)$.
\par
\begin{figure*}[!ht]
\centering
\setlength{\abovecaptionskip}{-0.2cm}
\includegraphics[width=11cm]{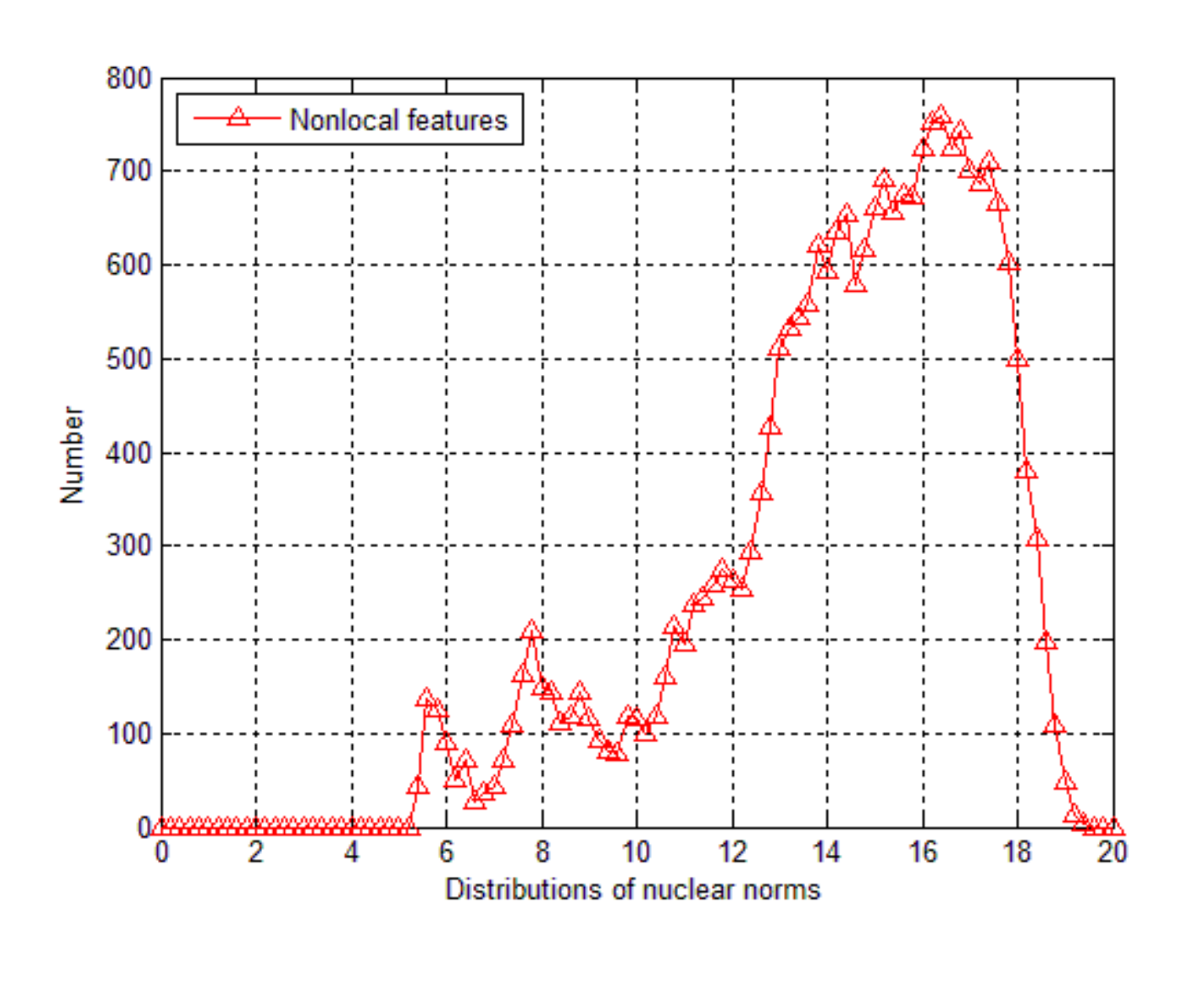}
\caption{Nuclear norm distributions of the matrices consisting of nonlinearly mapped nonlocal feature vectors. The feature vectors and their nonlocal neighbors are concatenated to form matrices of nonlocal features.}
\label{fig:NonlocalLowrank}
\end{figure*}

In Fig. \ref{fig:NonlocalLowrank}, we draw the nuclear norm distributions of the nonlocal matrices consisting of nonlinearly mapped nonlocal feature vectors. The feature vectors extracted from a test image and their $20$ nonlocal neighbors are concatenated to form matrices of nonlocal features. Note that the nuclear norm of nonlinear mapped nonlocal matrix $\phi(Y)$ is calculated as $||\phi(Y)||_*=tr(\Sigma(\phi(Y)^T\phi(Y)))=tr(\Sigma(K_{Y,Y}))$ where $\Sigma(K_{Y,Y})$ denotes the diagonal matrix with the eigenvalues of $K_{Y,Y}$ on the diagonal and ${tr(\cdot )}$ is the matrix trace operator. It shows that the matrices of nonlocal features tend to have relatively lower nuclear norms than their maximum ($21$), which indicates the low-rank property of nonlocal matrices.

\par Thus, with this preservation of low-rank property, the optimization problem of (\ref{ref:eqOrgProblem}) in kernel space can be rewritten as:
\begin{equation}
\label{ref:eqOrgProblem1}
W_j = \mathop{\rm{argmin}}\limits_{W_j} \frac{{\rm{1}}}{{\rm{2}}}{\rm{||}}{\phi(Y_j^A)} - {\phi(Y_j^D)}{W_j}{\rm{||}}_{\rm{F}}^{\rm{2}} + {\rm{\lambda}}_{\rm{1}}  {\rm{||}}{W_j}{\rm{||}}_{\rm{1}} + {\rm{\lambda}}_{\rm{2}} {\rm{||}}{W_j}{\rm{||}}_{\rm{*}}
\end{equation}
\par However, since the optimization problems of nuclear norm and $\ell 1$ norm (\ref{ref:eqOrgProblem1}) are difficult to solve simultaneously, we introduce two more relaxation variables and impose fidelity constraints between the pairs of relaxation variables:
\multlinegap=0pt
\begin{multline}
\label{ref:eqRlxProblem}
W_{{1\sim3},j} = \mathop{\rm{argmin}}\limits_{{W}_{{1\sim3},j}} \frac{{\rm{1}}}{{\rm{2}}}{\rm{||}}{\phi(Y_j^A)} - \phi(Y_j^D) {W}_{3,j} {\rm{||}}_{\rm{F}}^{\rm{2}} + {{\rm{\lambda }}_{\rm{1}}}{\rm{||}}W_{1,j}{\rm{|}}{{\rm{|}}_{\rm{1}}} + {{\rm{\lambda }}_{\rm{2}}}{\rm{||}}W_{2,j}{\rm{|}}{{\rm{|}}_{\rm{*}}},
\\{\rm{s.t.}}W_{3,j} = W_{1,j};W_{3,j} = W_{2,j}
\end{multline}
We use inexact augmented Lagrange multiplier (IALM) method \cite{Lin2009} to solve problem (\ref{ref:eqRlxProblem}), which has also been used to efficiently solve other low-rank problems (i.e. RPCA for low rank matrix recovery \cite{Candes2011}). We add two more variables to relax the fidelity constraints. The augmented Lagrange function for (\ref{ref:eqRlxProblem}) is:
\multlinegap=0pt
\begin{multline}
\label{ref:eqTerProblem}
    L (W_{{1\sim3},j}) = \frac{\rm{1}}{\rm{2}}  {\rm{||}} \phi(Y_j^A) - \phi(Y_j^D) W_{3,j} {\rm{||}}_{\rm{F}}^{\rm{2}}  + {\mathbf{\lambda }}_{\rm{1}} {\rm{||}} W_{1,j} {\rm{||}}_{\rm{1}} + {\mathbf{\lambda}}_{\rm{2}} {\rm{||}} W_{2,j}{\rm{||}}_{\rm{*}}
\\ + tr( \Lambda_1^T ( W_{1,j} - W_{3,j} )) + tr( \Lambda _2^T ( W_{2,j} - W_{3,j} ))
\\ + \frac{{\rm{u_1}}}{{\rm{2}}}  {\rm{||}} W_{1,j} - W_{3,j} {\rm{||}}_{\rm{F}}^{\rm{2}} + \frac{{{{\rm{u}}_{\rm{2}}}}}{{\rm{2}}}{\rm{||}} W_{2,j} - W_{3,j} {\rm{||}}_{\rm{F}}^{\rm{2}}
\end{multline}
where ${tr(\cdot )}$ is the operator to get matrix trace. $\lambda_{\rm{1}}$ and $\lambda_{\rm{2}}$ are scalar constants. $\Lambda_{\rm{1}}$ and $\Lambda_{\rm{2}}$ are Lagrange multiplier matrices. ${{\rm{u}}_{\rm{1}}}$ and ${{\rm{u}}_{\rm{2}}}$ are the parameters balancing the difference between pairs of objective variables and other regularization terms.

\subsubsection{Optimization of KLRSC}
There are three objective variables $W_{{1\sim3},j}$ in (\ref{ref:eqTerProblem}) which we alternatively update, followed by the adjustment of multipliers. Soft-threshold operations on matrix elements and singular values are used to solve the problem of ${\ell 1}$-norm and nuclear norm optimizations. The update steps of $W_{{1\sim3},j}$ and the multipliers are given below.

\noindent \textbf{Update $\bm{W_{1,j}}$}
\par
Firstly, we update $W_{1,j}$ and meanwhile fix other variables. The optimization function with respect to $W_{1,j}$ derived from (\ref{ref:eqTerProblem}) can be formulated as:
\begin{equation}
\label{ref:1}
\arg \mathop {\min }\limits_{W_{1,j}} \frac{\lambda_1}{u_1}{||}W_{1,j}{||}_1 + \frac{1}{2}||W_{1,j}-(W_{3,j}+\frac{1}{u_1}\Lambda_1)||_F^2
\end{equation}
The ${\ell 1}$-norm optimization problem of (\ref{ref:1}) can be solved by soft-thresholding:
\begin{equation}
\label{ref:11}
 W_{1,j} = S_{\frac{{{\lambda _1}}}{{{u_1}}}}({W_{3,j}} + \frac{1}{{{u_1}}}{\Lambda_1})
\end{equation}
where $S_\lambda(W)=$sign$(W)$max$(0,|W|-\lambda)$ is a shrinkage operator on values of matrix $W$.

\noindent \textbf{Update $\bm{W_{2,j}}$}
\par
Then we update $W_{2,j}$ and fix others by solving the following optimization problem:
\begin{equation}
\label{ref:2}
\arg \mathop {\min }\limits_{W_{2,j}} \frac{\lambda_2}{u_2}{||}W_{2,j}{||}_* + \frac{1}{2}||W_{2,j}-(W_{3,j}+\frac{1}{u_2}\Lambda_2)||_F^2
\end{equation}
The nuclear norm optimization problem of (\ref{ref:2}) can be solved by singular value soft-thresholding:
\begin{equation}
\label{ref:21}
 W_{2,j} = \Im_{\frac{{{\lambda _2}}}{{{u_2}}}}({W_{3,j}} + \frac{1}{{{u_2}}}{\Lambda_2})
\end{equation}
where $\Im_\lambda(W)=U_WS_\lambda(\Sigma_W)V_W^T$ is a shrinkage operator on singular values of matrix $W$ and $U_{W}\Sigma_{W}V_W^T$ is the singular value decomposition of $W$.

\noindent \textbf{Update $\bm{W_{3,j}}$}
\par
The optimization function with respect to $W_{3,j}$ is given by:
\multlinegap=0pt
\begin{multline}
\label{ref:3}
         \arg \mathop {\min }\limits_{{W_{3,j}}} \frac{{\rm{1}}}{{\rm{2}}}{\rm{||}} \phi(Y_j^A){\rm{ - }} \phi( Y_j^D){W_{3,j}}{\rm{||}}_{\rm{F}}^{\rm{2}} + tr(\Lambda_1^T({W_{3,j}} - {W_{1,j}}))
           \\+tr(\Lambda_2^T({W_{3,j}} - {W_{2,j}}))+ \frac{{{u_1}}}{{\rm{2}}}||{W_{3,j}} - {W_{1,j}}||_F^2 + \frac{{{u_2}}}{{\rm{2}}}||{W_{3,j}} - {W_{2,j}}||_F^2
\end{multline}
Solving the optimization problem (\ref{ref:3}), we update $W_{3,j}$ by:
\begin{equation}
    \begin{aligned}
    W_{3,j} = & (\phi (Y_j^D)^T\phi(Y_j^D) + (u_1+ u_2){I})^{ - 1}Z\\
    &=(K_{{Y_j^D}{Y_j^D}} - (u_1+ u_2){I})^{ - 1}Z
    \end{aligned}
\end{equation}
where $I$ is an identity matrix and the matrix $Z$ is represented as:
\begin{equation}
    \begin{aligned}
    Z=& \phi (Y_j^D)^T\phi(Y_j^A) + u_1{W_{1,j}} - \Lambda_1 + u_2{W_{2,j}} - {\Lambda_2}\\
    &=K_{{Y_j^D}{Y_j^A}} - u_1{W_{1,j}} + \Lambda_1 - u_2{W_{2,j}} + {\Lambda_2}
    \end{aligned}
\end{equation}

\noindent \textbf{Update multipliers}
\begin{equation}
\begin{aligned}
{\Lambda_1} = {\Lambda_1} + {u_1}&({W_{3,j}} - {W_{1,j}});{\Lambda_2}{= }{\Lambda_2} + {u_2}({W_{3,j}} - {W_{2,j}})\\
&{u_1} = \rho {u_1};{u_2} = \rho {u_2}
\end{aligned}
\end{equation}
where $\rho>1$ is a scalar constant.
\par When the changes of objective variables during updates are below a defined threshold $e$, the optimization reaches convergence. We summarize this optimization in Algorithm \ref{tab:KLRSC}.
\begin{algorithm}
\caption{Optimization of Kernel Based Low-rank Sparse Coding Problem}
\begin{algorithmic}[1]
\REQUIRE Data $Y_j^A$, Sub-dictionary $Y_j^D$ and parameters $\lambda_{\rm{1}}$, $\lambda_{\rm{2}}$, $\Lambda_{\rm{1}}$, $\Lambda_{\rm{2}}$, ${{\rm{u}}_{\rm{1}}}$ and ${{\rm{u}}_{\rm{2}}}$\\
\textbf{while} not converged \textbf{do}:\\
 \quad \textbf{Fix the others and update} $W_{1,j}$ \\
    \quad ${ W_{1,j}} = {S_{\frac{{{\lambda _1}}}{{{u_1}}}}}({W_{3,j}} + \frac{1}{{{u_1}}}{\Lambda_1})$\\
 \quad\textbf{Fix the others and update} $W_{2,j}$ \\
    \quad${W_{2,j}} = {\Im _{\frac{{{\lambda _2}}}{{{u_2}}}}}(W_{3,j} + \frac{1}{{{u_2}}}{\Lambda_2})$\\
 \quad\textbf{Fix the others and update} $W_{3,j}$ \\
     \quad${W_{3,j}}={(K_{{Y_j^D}{Y_j^D}} - {u_1}{\rm{I - }}{u_2}{\rm{I}})^{ - 1}}{\rm{(}}K_{{Y_j^D}{Y_j^A}} - {u_1}{W_{1,j}} + {\Lambda_1} - {u_2}{W_{2,j}} + {\Lambda_2})$\\
 \quad\textbf{Fix the others and update the multipliers} $\Lambda_{\rm{1}}$,$\Lambda_{\rm{2}}$,${{\rm{u}}_{\rm{1}}}$ and ${{\rm{u}}_{\rm{2}}}$\\
    \quad  ${\Lambda_1}{\rm{ = }}{\Lambda_1} + {u_1}({W_{3,j}} - {W_{1,j}});{\Lambda_2}{\rm{ = }}{\Lambda_2} + {u_2}({W_{3,j}} - {W_{2,j}})$\\
    \quad  ${u_1} = \rho {u_1};{u_2} = \rho {u_2}$\\
\textbf{end while}
\ENSURE $W_{3,j}$
\end{algorithmic}
\label{tab:KLRSC}
\end{algorithm}

\subsubsection{Effectiveness of KLRSC}
\begin{figure*}[!ht]
\centering
\setlength{\abovecaptionskip}{-0.5cm}
\includegraphics[width=10cm]{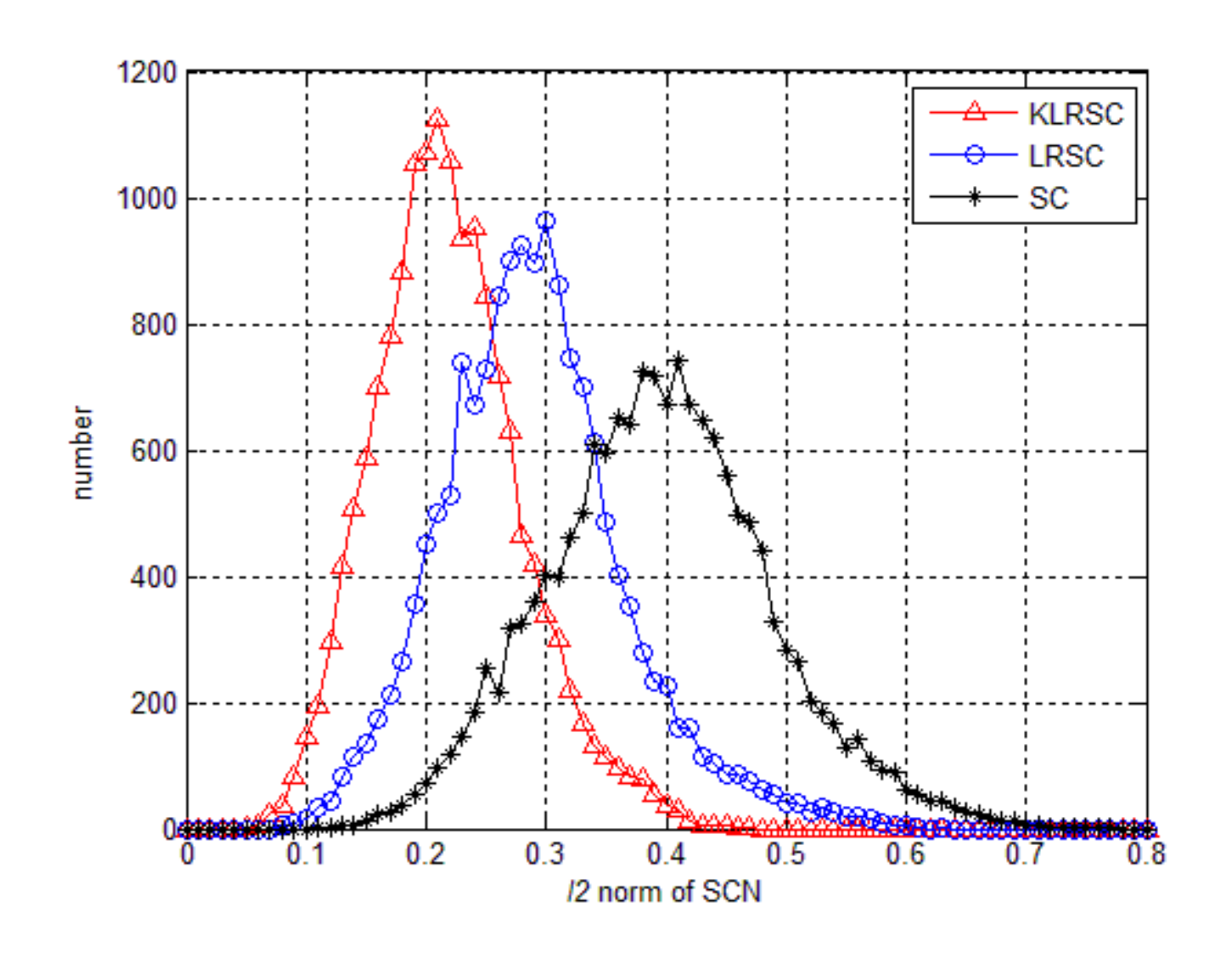}
\caption{${\ell 2}$ norm distributions of SCN for KLRSC, LRSC and SC.}
\label{fig:SCN}
\end{figure*}

To explain the effectiveness of KLRSC, we perform an experiment to investigate the statistical property of sparse coding noises (SCN) for different coding methods. We use $lena$ image as a test image. Its LR counterpart is generated through blurring (Gaussian kernel with standard deviation $1.6$), down-sampling and up-scaling (with a factor of $1.25$). We collect $15625$ pairs of LR and HR features from the LR and HR images. DCT dictionary is used in our experiment. We denote method of sparse coding with low-rank constraint as 'LRSC' which appeared in our preliminary work \cite{prework}. We firstly calculate the sparse coefficients for them using KLRSC, LRSC and SC, respectively. We calculate SCN by following the definition in \cite{Dong2013}. In our experiment, We evaluate SCN by ${\ell 2}$ norm. In Fig \ref{fig:SCN}, we draw the ${\ell 2}$ norm distributions of SCN for KLRSC, LRSC and conventional SC. The distribution for KLRSC, LRSC and SC is drawn in red, blue and black lines, respectively. It is shown that KLRSC get lower SCN than the other two methods do, which means that the proposed KLRSC approach effectively suppress SCN utilizing the low-rank property of nonlocal-similarity and improve the coding performance.

\subsubsection{Post-processing procedure}
\par
 When the optimization converges, the solution $W_{3,j}$ becomes both sparse and low-rank. Then we distil the first column of $W_{3,j}$ as the sparse weight $w_j^{lr}$ for the reconstruction of the HR patch $x_j$ because low-rank constraint does not change the identities of columns. The HR patch $x_j$ can be sparsely represented upon $X_j^D$ as:
\begin{equation}
\label{ref:eqHreconstruction}
x_j = X_j^D w_j^{lr} \cdot ||l_j||_2 + \bar l
\end{equation}
where $||l_j||_2$ is the ${\ell 2}$-norm of the corresponding LR feature and $\bar l$ denotes the average intensity of the corresponding LR patch. Having obtained all HR patches $X_i = \{ x_j \}_i$, we merge them into the layer $I_i$ by averaging the intensity of the overlapping pixels between the adjacent patches.
\par
To enhance quality of the reconstructed interim layer, we apply IBP algorithm \cite{Irani} and pixel-wise autoregressive (AR) model \cite{Dong2011} to both enforce the global reconstruction constraint between the interim layer and the input $I_0$ and refine the relation between neighboring pixels.
\par
The \emph{j}-th pixel of the reconstructed layer is expected to be predicted as a linear combination of its neighboring pixels in a $3\times3$ square window: $s_j=a_j^Tq_j$, where $s_j$ is the central pixel and $q_j$ is the vector consisting of its neighbors. To learn the combination weights $a_j$, we collect the \emph{N} nearest neighbors of the $s_j$ centered patch from other already reconstructed HR layers. These patches are assumed to share the same neighboring relationship. The combination weights $a_j$ can be obtained by the following optimization problem:


\begin{equation}
a_j=\arg \mathop {\min }\limits_{a_j}\sum_{n=1}^N(s_j^n-a_j^Tq_j^n)^2+\eta||a_j||_2^2
\end{equation}
$a_j$ can be derived by:
\begin{equation}
a_j=(QQ^T+\eta E)^{-1}QS^T
\end{equation}
where $Q=[q_j^1,q_j^2,...,q_j^N]$, $S=[s_j^1,s_j^2,...,s_j^N]$ and $E$ is the identity matrix. Thus, we regularize the estimated layer by minimize the AR prediction error and the global reconstruction error by:
\begin{equation}
I_i^{*}=\arg \mathop {\min }\limits_{I_i}||I_0-D_iB_iI_i||_F^2+\alpha||I_i-A_iI_i||_F^2+\beta||I_i-I_{i,0}||_F^2
\end{equation}
where $A_i$ describes pixel-wise relationships in $I_i$, $I_{i,0}$ denotes the initial HR estimation, $I_0$ is the LR observation, $D_i$ and $B_i$ are the down-sampling and blurring operator of the $i$-th layer, respectively. The layer $I_{i}$ is updated by:
\begin{equation}
\label{ref:AR}
I_{i,t+1}=I_{i,t}+\tau [B_i^T D_i^T (I_0-D_iB_iI_{i,t})-\alpha(E-A_i)^T(E-A_i)I_{i,t}-\beta (I_{i,t}-I_{i,0})]
\end{equation}
where $\tau$ is the step size for gradient descent.
\par According to the self-similarity learning framework, we repeat the aforementioned low-rank sparse representation based SR for ${N_U}$ times followed by a fine adjustment to get the final SR result.

\subsection{Summary}
The complete SR process is summarized in Algorithm \ref{tab:TableAl1}.
\begin{algorithm}[!ht]
\caption{Proposed Kernel Based Low-Rank Sparse Model for Single Image Super-Resolution}
\begin{algorithmic}[1]
\REQUIRE LR image $I_L(I_0)$ and up-scaling factor $p$\\
\STATE \textbf{Initialization}\\
    Set the input image $I_L$ as initial layer $I_0$. Create the double pyramids from the input image using (\ref{ref:HR_pyramid}) and (\ref{ref:LR_pyramid}). Collect self-examples from the pyramids to generate the LR dictionary $L$ and the HR dictionary $H$\\
\STATE\textbf{Upscaling} \\
    Gradual magnification loop:\\
        \qquad \textbf{for} {$i=1$ to $N_U$} do\\
            \qquad\qquad 1) Enlarge the last layer $I_{i-1}$ by a factor of $s$ to build the layer \\
            \qquad\qquad \quad $U(I_i) \leftarrow (I_{i-1})_{\uparrow s}$.\\
            \qquad\qquad  2) Partition the layer $U(I_i)$ into LR patches $Y_i=\{y_j\}_i$.\\
            \qquad\qquad  3) Compute the kernel based low-rank sparse representation \\
            \qquad\qquad \quad coefficients of each LR patch $y_j$.\\
            \qquad\qquad  4) Reconstruct the HR patch $x_j$\\
            \qquad\qquad  5) Merge the HR patches $\{x_j\}_i$ into the layer $I_i$.\\
            \qquad\qquad  6) Refine the layer $I_i$ by IBP and AR using (\ref{ref:AR}).\\
         \qquad \textbf{end for}\\
\STATE \textbf{Image size adjustment}\\
    Down-sample the final layer $I_{N_U}$ to get $I_H$.
\ENSURE HR image $I_H$\\
\end{algorithmic}
\label{tab:TableAl1}
\end{algorithm}

\section{Experimental results}
\begin{figure*}[!ht]
\centering
\setlength{\abovecaptionskip}{-1cm}
\includegraphics[width=\textwidth]{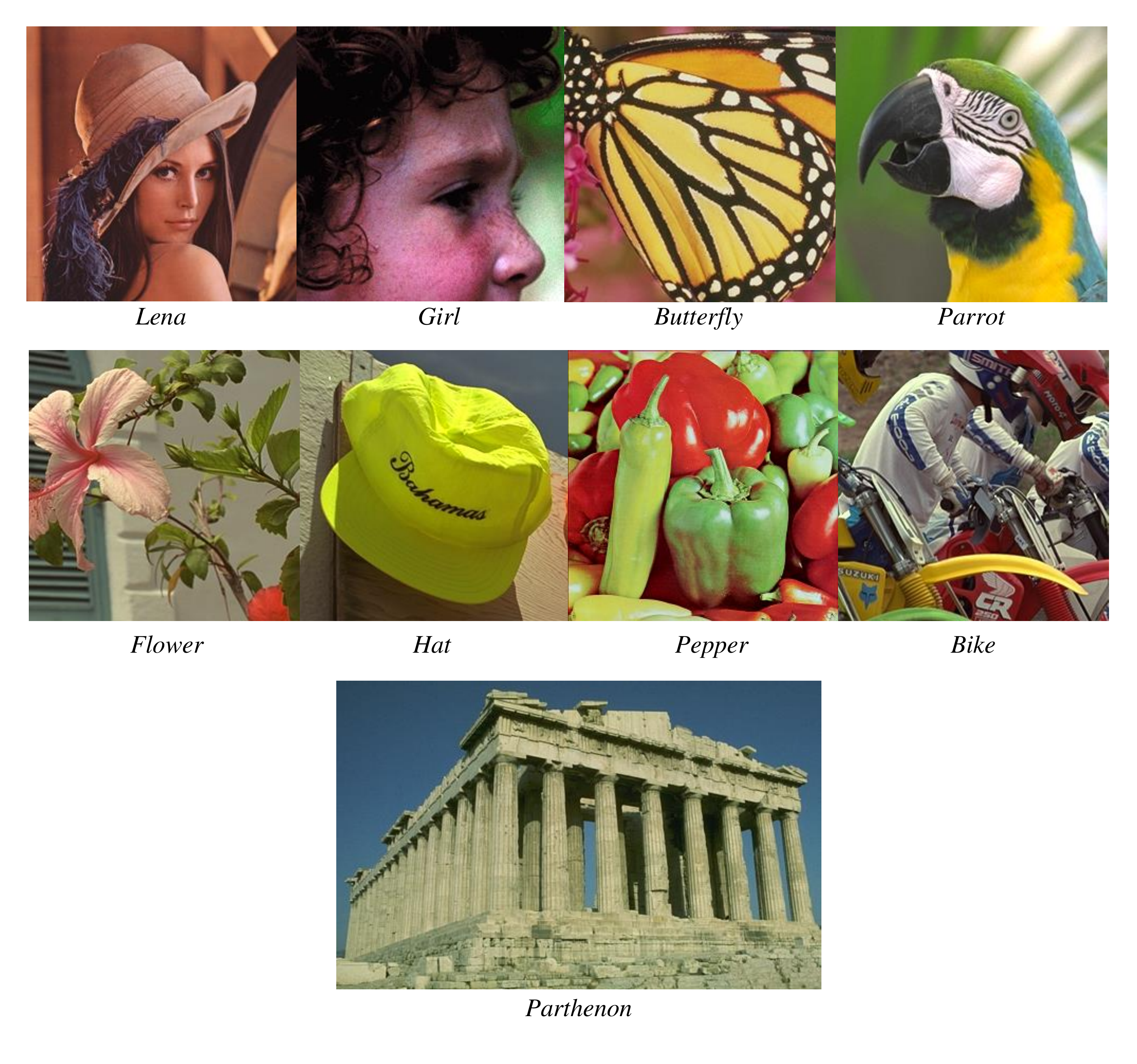}
\caption{Test Images}
\label{fig:TestImages}
\end{figure*}

In our experiments, we use nine test images from the software package for \cite{Dong2011}. These images (see Fig. \ref{fig:TestImages}) cover various contents including humans, animals, plants and man-made objects. The size of image \emph{parthenon} is $459\times292$ and the size of other images is $256\times256$. We compare our method with SC-SR \cite{Yang2010}, ASDS \cite{Dong2011}, LRNE-SR \cite{Chen2014}, DM-SR \cite{Bevilacqua2014}, NCSR \cite{Dong2013} and Aplus \cite{Aplus2015}. Since the human visual systems are more sensitive to luminance changes in color images, we only perform our proposed method on the luminance component. The SR performances are evaluated in the luminance channel by the peak signal-to-noise ratio (PSNR) and the structural similarity (SSIM) \cite{SSIM} objectively.

\subsection{Experimental setting}
\par
The color test images are blurred with $7 \times 7$ Gaussian kernel with standard variation $\sigma = 1.6$ and then down-sampled by bicubic interpolation to generate the LR input images. All the layers of images are split into $7 \times 7$ patches with overlap of five pixels. The number of layers of LR and HR pyramids to train the self-example dictionary is $N_D=4$. The up-scaling factor for each time of magnification is $s = 1.25$. The standard variance of Gaussian Blur kernel for the generation of the $n$-th layer $I_{-n}$ is computed by (\ref{ref:Blur_kernel_variance}). The number of the neighbors for low-rank sparse representation ${K_D}$ is 60. The number of similar nonlocal neighbors ${K_N}$ is 20. We set $\lambda_1=0.07$, $\lambda_2=0.07$, $\rho=1.5$. We obtain initialization of codes $Z_0$ by standard sparse coding \cite{Yang2010} and let $Z_1=Z_2=Z_3=Z_0$ at the beginning of optimization.  We set $u_1=1/{max(\Sigma(Z_0))}$ and $u_2=1/||Z_0||_\infty$. For the stage of IBP and AR regularization, the window size is $3\times3$. The maximum iteration times is set to $300$. We set the step size $\tau=0.5$. The parameters $\alpha$ and $\beta$ are set to $0.05$ and $0.01$.
\par
For fairness of the comparisons, according to the experimental setting, we retrain the LR-HR dictionary for \cite{Yang2010} and \cite{Chen2014} and change all down-sampling and up-scaling for \cite{Dong2013} and \cite{Dong2011} to bicubic and retrain its AR models and nonlocal adaptive regularization models before implementation.

\subsection{Experimental results}
\par The PSNRs and SSIMs of different methods for comparisons are shown in Table \ref{tab:Table3} for the scaling factor $p=3$ and Table \ref{tab:Table4} for the scaling factor $p=4$, respectively. Our proposed method gets better quantitative SR performances on most of the test images than other methods. The average gains of our proposed method for the scaling factor $p=3$ over the second best method are 0.293dB in PSNR and 0.0078 in SSIM. In the case of $p=4$, the average gains are 0.420dB in PSNR and 0.0117 in SSIM.
\begin{table}[!ht]
  \centering
  \caption{PSNRs (dB) and SSIMs by different SR methods with the scaling factor $p=3$ ($\sigma=1.6$)}
  \footnotesize
       \begin{tabular}{p{0.92cm}<{\centering}p{0.9cm}<{\centering}p{0.85cm}<{\centering}p{0.85cm}<{\centering}p{1.03cm}<{\centering}p{0.85cm}<{\centering}p{0.85cm}<{\centering}p{0.95cm}<{\centering}p{1.05cm}<{\centering}}
    \toprule
 \multirow{2}[2]{*}{Image}&{Bi-}&{SC-}&ASDS&LRNE-&DM-&NCSR&Aplus&{Pro-}\\
          &{cubic}&SR\cite{Yang2010}&\cite{Dong2011}&SR\cite{Chen2014}&SR\cite{Bevilacqua2014}&\cite{Dong2013}&\cite{Aplus2015}&{posed}\\
    \toprule
    \multirow{2}[2]{*}{Lena}&29.600&30.489&31.232 &30.618&30.691&31.357 &31.610&\textbf{31.884}\\
          &0.8306&0.8556& 0.8693 &0.8523&0.8555&0.8747 &0.8708&\textbf{0.8791}\\
          \midrule
    \multirow{2}[2]{*}{Girl}  &32.724 & 33.529& 33.753& 33.422 & 33.599 &33.982&33.367&\textbf{34.319}\\
          &0.8162&0.8386& 0.8417& 0.8321 & 0.8410 & 0.8489 &0.8205&\textbf{0.8540}\\
          \midrule
   {Butt-}   & 23.103& 24.337&25.196 & 24.815 & 24.940 &25.391 &\textbf{26.808}&26.735\\
   {erfly}     &0.7926 & 0.8375 &0.8680& 0.8553 & 0.8615 & 0.8754&0.8980&\textbf{0.9003}\\
           \midrule
    \multirow{2}[2]{*}{Parrot} &27.406& 28.488 &29.309 & 28.339 &28.728 &29.355 &29.460&\textbf{29.503}\\
           &0.8692& 0.8910&0.9025& 0.8876 & 0.8932 & 0.9063&0.9049 &\textbf{0.9087}\\
          \midrule
    \multirow{2}[2]{*}{Flower}&26.935& 27.949 &28.489 & 27.885 &28.177 &28.581&28.802&\textbf{29.230}\\
          &0.7594& 0.8025& 0.8215  & 0.7978 & 0.8113 & 0.8285& 0.8382&\textbf{0.8437}\\
           \midrule
    \multirow{2}[2]{*}{Pepper} &27.269& 28.275& 29.065 & 28.605 &28.527 &29.140 &29.514 &\textbf{30.059}\\
           &0.8393& 0.8628&0.8827 & 0.8697 & 0.8704 &0.8862  &0.8764&\textbf{0.8967}\\
          \midrule
    \multirow{2}[2]{*}{Bike} & 22.502& 23.385& 23.928 & 23.407 & 23.555 & 24.010 &24.264&\textbf{24.429}\\
           &0.6640& 0.7235&0.7494& 0.7177 & 0.7313 &0.7563  &\textbf{0.7782}&0.7721\\
           \midrule
    \multirow{2}[2]{*}{Hat} &28.958& 29.824& 30.363& 30.030 & 30.107 &30.530 &30.915 &\textbf{31.177}\\
    &0.8233& 0.8454 & 0.8568& 0.8508 & 0.8536 & 0.8641 &0.8670&\textbf{0.8739}\\
           \midrule
    {Parth-} &25.593& 26.081& 26.543 & 26.137 & 26.228 & 26.608&26.823&\textbf{26.869}\\
    {enon} &0.6683& 0.6978& 0.7118& 0.6923 & 0.6992  & 0.7180&\textbf{0.7315} &0.7266\\
     \toprule
    \multirow{2}[2]{*}{Avg.}  &27.121& 28.040&28.653& 28.140 & 28.284 & 28.773 &29.063&\textbf{29.356}\\
          &0.7848& 0.8172& 0.8337& 0.8173 & 0.8241 & 0.8398 &0.8428&\textbf{0.8506}\\
    \bottomrule
    \end{tabular}%
 \label{tab:Table3}%
\end{table}
\begin{table}[!ht]
  \centering
   \caption{PSNRs (dB) and SSIMs by different SR methods with the scaling factor $p=4$ ($\sigma=1.6$)}
  \footnotesize
    \begin{tabular}{p{0.92cm}<{\centering}p{0.9cm}<{\centering}p{0.85cm}<{\centering}p{0.85cm}<{\centering}p{1.03cm}<{\centering}p{0.85cm}<{\centering}p{0.95cm}<{\centering}p{0.85cm}<{\centering}p{1.05cm}<{\centering}}

    \toprule
\multirow{2}[2]{*}{Image}&{Bi-}&{SC-}&ASDS&LRNE-&DM-&NCSR&Aplus&{Pro-}\\
         &{cubic}&SR\cite{Yang2010}&\cite{Dong2011}&SR\cite{Chen2014}&SR\cite{Bevilacqua2014}&\cite{Dong2013}&\cite{Aplus2015}&{posed}\\
 \toprule
      \multirow{2}[2]{*}{Lena}&27.820& 28.529 & 29.254 & 28.591 & 29.039 &29.908 & 29.610 & \textbf{30.083} \\
         &0.7605& 0.7849 & 0.8047 & 0.7838 & 0.8000 & 0.8234& 0.8204 & \textbf{0.8252} \\
  \midrule
      \multirow{2}[2]{*}{Girl}  &31.263& 31.814 & 32.211 & 31.826 & 32.211 &32.109 & 31.971 & \textbf{32.635} \\
          &0.7601& 0.7768 & 0.7873 & 0.7743 & 0.7885 & 0.7751 & 0.7699 & \textbf{0.7978} \\
  \midrule
   {Butt-}  &21.438& 22.656 & 23.666 & 22.737 & 23.781 & 24.208 & 23.996 & \textbf{24.827} \\
          {erfly} &0.7121& 0.7555 & 0.8145 & 0.7721 & 0.8198 & 0.8295 & 0.8316 & \textbf{0.8511} \\
  \midrule
   \multirow{2}[2]{*}{Parrot} &25.512& 26.416 & 26.966 & 26.276 & 26.796 & \textbf{27.428}& 27.228 & 27.183 \\
          &0.8145& 0.8380 & 0.8521 & 0.8346 & 0.8486 & \textbf{0.8654}& 0.8638 & 0.8639 \\
  \midrule
     \multirow{2}[2]{*}{Flower} &25.259& 26.113 & 26.452 & 25.952 & 26.510 & 26.850 & 26.706  & \textbf{27.022} \\
    &0.6689& 0.7157 & 0.7306 & 0.7071 & 0.7366 & 0.7552 & 0.7498  & \textbf{0.7597} \\
  \midrule
     \multirow{2}[2]{*}{Pepper} &25.527& 26.358 & 27.153 & 26.480 & 27.011 & 27.668 & 27.367 & \textbf{27.887} \\
          &0.7660& 0.7866 & 0.8174 & 0.7925 & 0.8135 & 0.8296 & 0.8279 & \textbf{0.8340} \\
  \midrule
\multirow{2}[2]{*}{Bike}  &21.021& 21.747 & 22.248 & 21.738 & 22.172 & 22.646 & 22.423  & \textbf{22.716} \\
        &0.5585& 0.6170 & 0.6488 & 0.6108 & 0.6462 & 0.6774& 0.6673  & \textbf{0.6800} \\
  \midrule
\multirow{2}[2]{*}{Hat}&27.487& 28.187 & 28.699 & 28.344 & 26.510 & 29.347 & 29.022 & \textbf{29.485} \\
&0.7771& 0.7933 & 0.8108 & 0.8009 & 0.7366 & 0.8277 & 0.8226 & \textbf{0.8319} \\
  \midrule

{Parth-}&24.875& 25.388 & 25.682 & 25.366 & 25.628 & 25.549 & 25.516 & \textbf{25.783} \\
 {enon}  &0.6296& 0.6574 & 0.6683 & 0.6518 & 0.6659 & 0.6620 & 0.6593 & \textbf{0.6748} \\
\toprule
\multirow{2}[2]{*}{Avg.}  &25.578& 26.356 & 26.926 & 26.368 & 26.629 & 27.301 & 27.093  & \textbf{27.513} \\
&0.7163& 0.7472 & 0.7705 & 0.7476 & 0.7617 &0.7828 & 0.7792 & \textbf{0.7909} \\
 \bottomrule
    \end{tabular}
 \label{tab:Table4}
\end{table}
\par
\begin{figure*}[!ht]
\setlength{\abovecaptionskip}{-0.5cm}
\includegraphics[width=\textwidth]{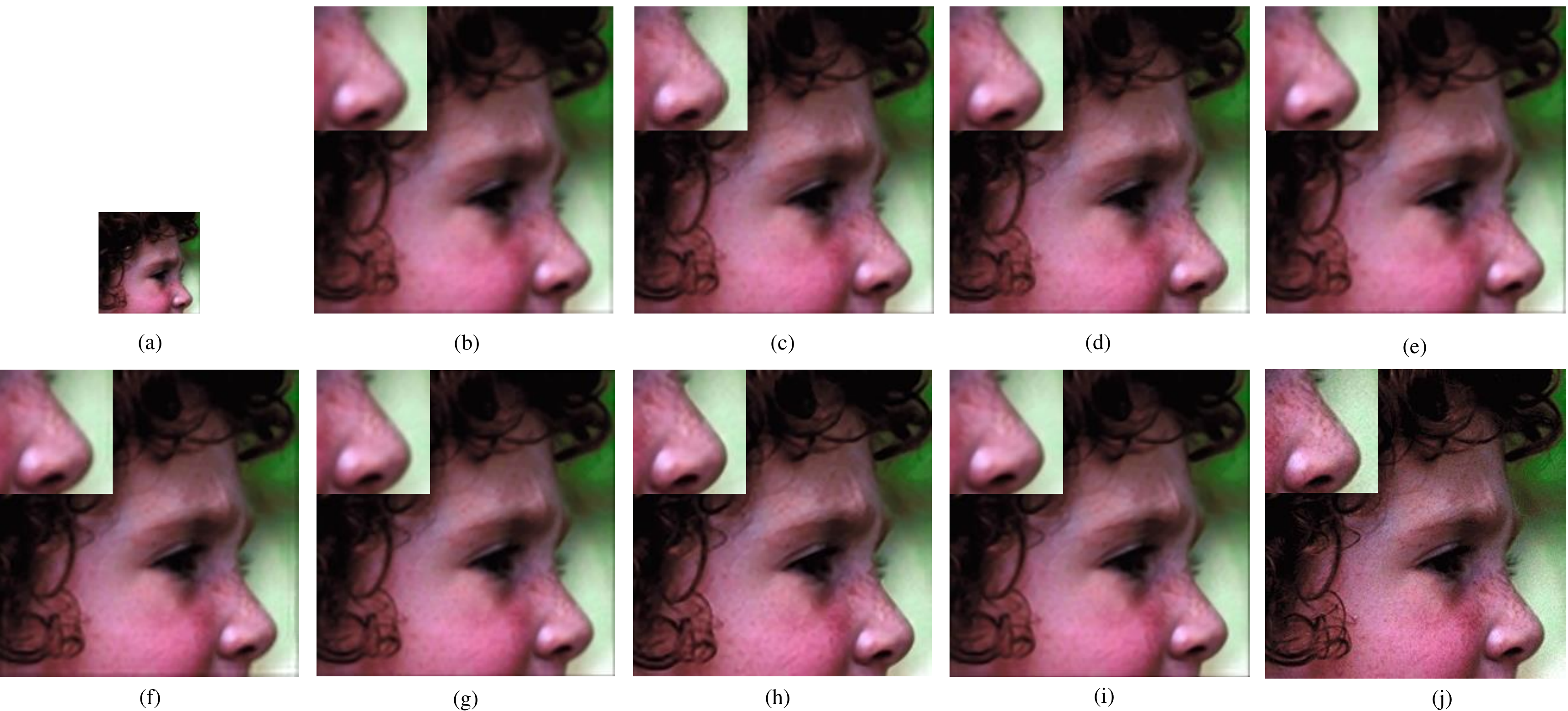}
\caption{Visual results comparison for image \emph{girl} ($\times3$, $\sigma=1.6$). (a)LR input. (b)bicubic. (c)SC-SR\cite{Yang2010}. (d)ASDS\cite{Dong2011}. (e)LRNE-SR\cite{Chen2014}. (f)DM-SR\cite{Bevilacqua2014}. (g)NCSR\cite{Dong2013}. (h)Aplus\cite{Aplus2015}. (i)proposed method. (j)ground truth.}
\label{fig:times3g}
\end{figure*}

\begin{figure*}[!htb]
\setlength{\abovecaptionskip}{-0.5cm}
\includegraphics[width=\textwidth]{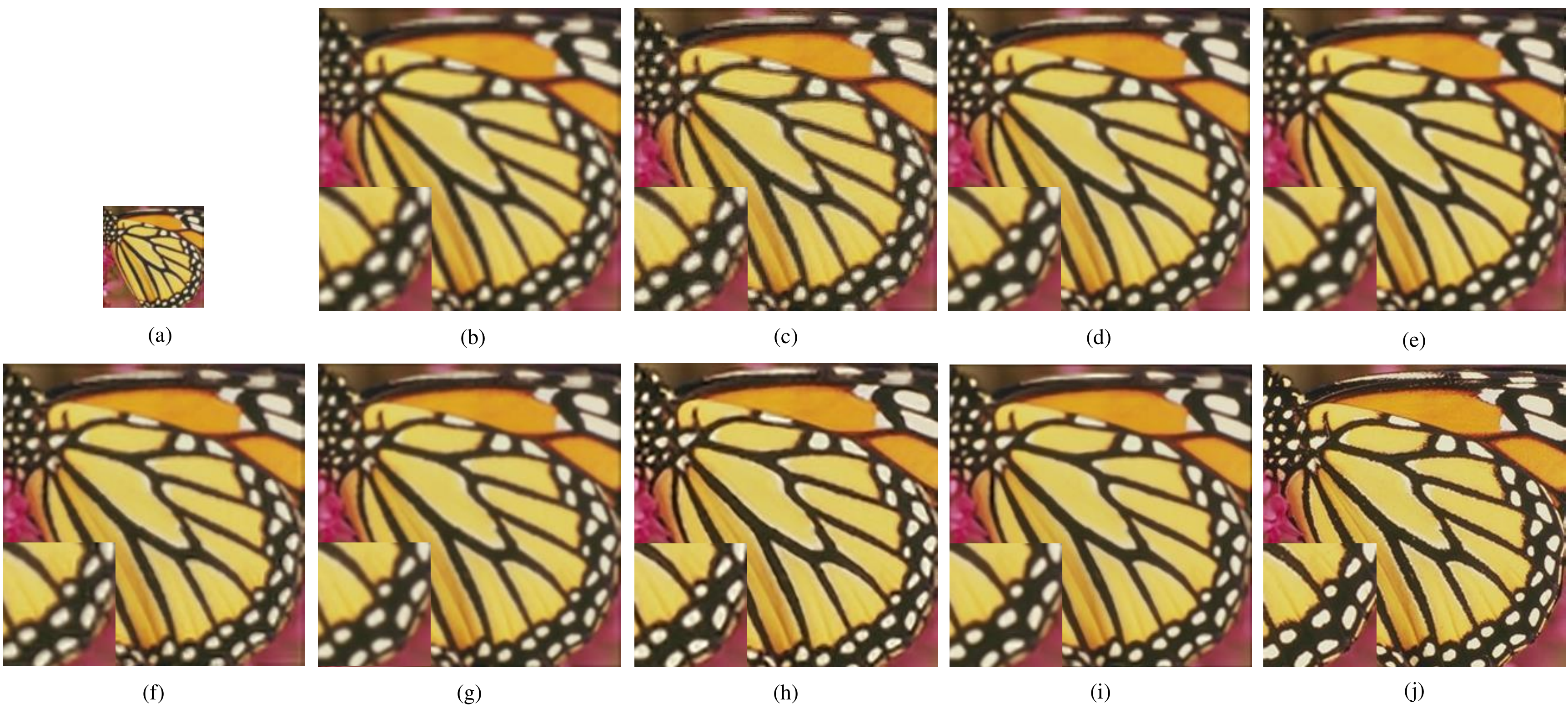}
\caption{Visual results comparison for image \emph{butterfly} ($\times3$, $\sigma=1.6$). (a)LR input. (b)bicubic. (c)SC-SR\cite{Yang2010}. (d)ASDS\cite{Dong2011}. (e)LRNE-SR\cite{Chen2014}. (f)DM-SR\cite{Bevilacqua2014}. (g)NCSR\cite{Dong2013}. (h)Aplus\cite{Aplus2015}. (i)proposed method. (j)ground truth.}
\label{fig:times3b}
\end{figure*}

\begin{figure*}[!htb]
\setlength{\abovecaptionskip}{-0.5cm}
\includegraphics[width=\textwidth]{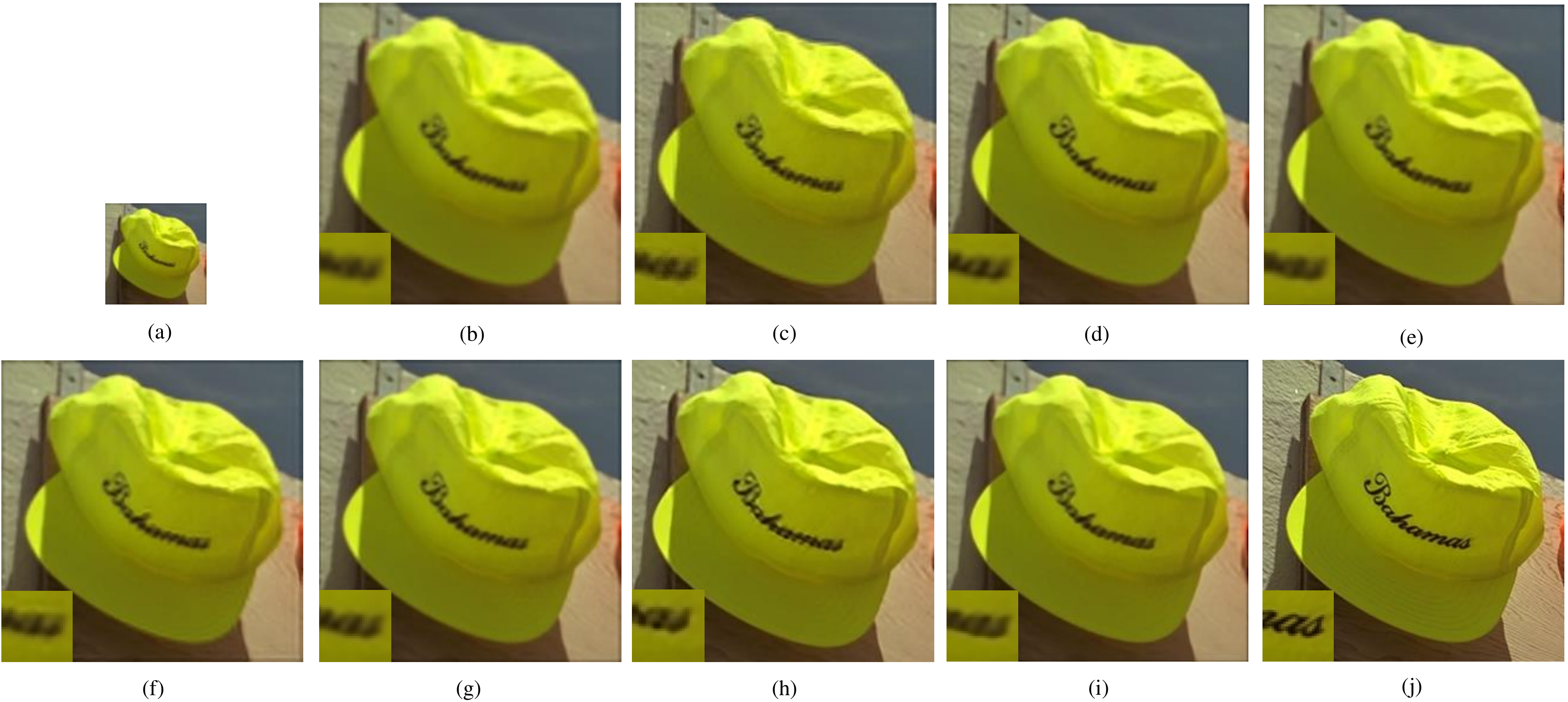}
\caption{Visual results comparison for image \emph{hat} ($\times3$, $\sigma=1.6$). (a)LR input. (b)bicubic. (c)SC-SR\cite{Yang2010}. (d)ASDS\cite{Dong2011}. (e)LRNE-SR\cite{Chen2014}. (f)DM-SR\cite{Bevilacqua2014}. (g)NCSR\cite{Dong2013}. (h)Aplus\cite{Aplus2015}. (i)proposed method. (j)ground truth.}
\label{fig:times3h}
\end{figure*}

\begin{figure*}[!htb]
\setlength{\abovecaptionskip}{-0.5cm}
\includegraphics[width=\textwidth]{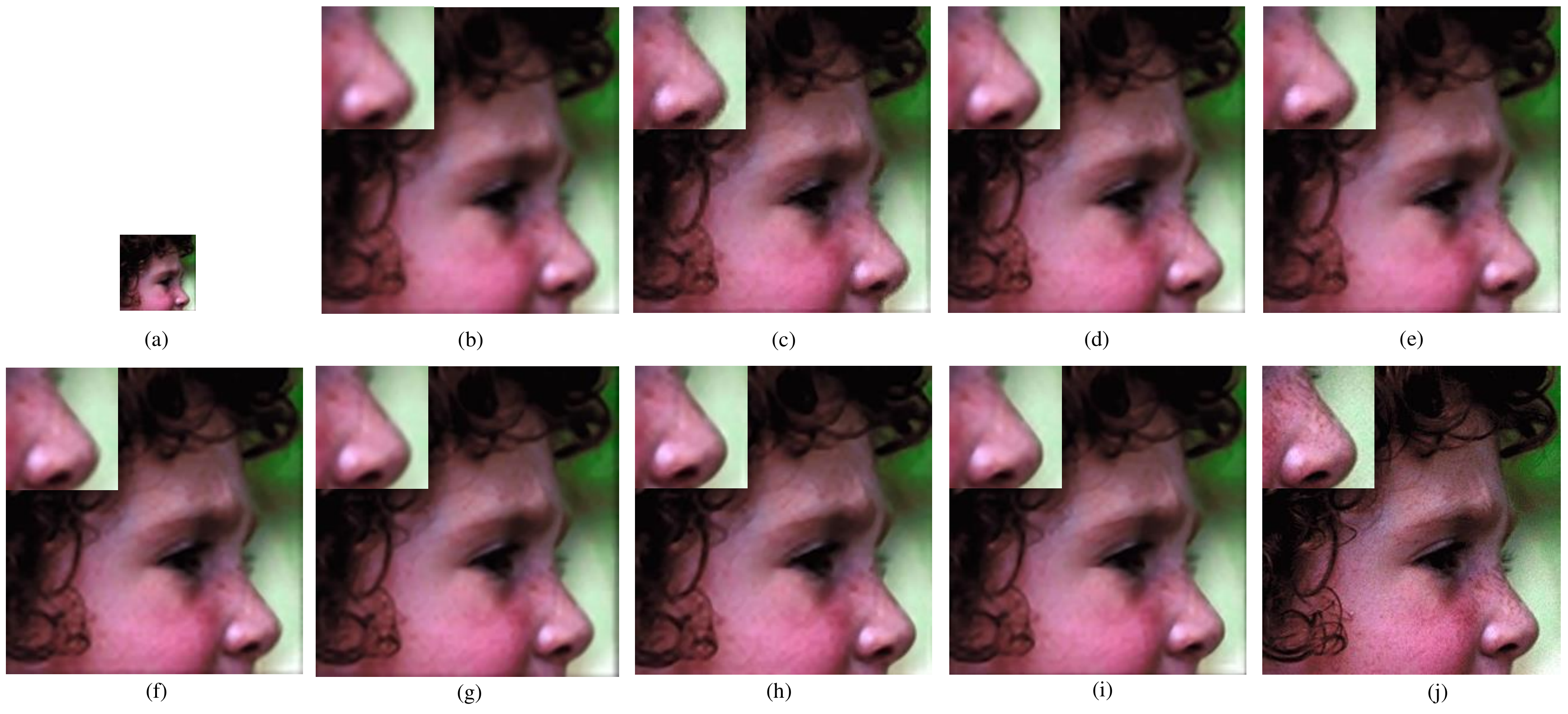}
\caption{Visual results comparison for image \emph{girl} ($\times4$, $\sigma=1.6$). (a)LR input. (b)bicubic. (c)SC-SR\cite{Yang2010}. (d)ASDS\cite{Dong2011}. (e)LRNE-SR\cite{Chen2014}. (f)DM-SR\cite{Bevilacqua2014}. (g)NCSR\cite{Dong2013}. (h)Aplus\cite{Aplus2015}. (i)proposed method. (j)ground truth.}
\label{fig:times4g}
\end{figure*}

\begin{figure*}[!htb]
\setlength{\abovecaptionskip}{-0.5cm}
\includegraphics[width=\textwidth]{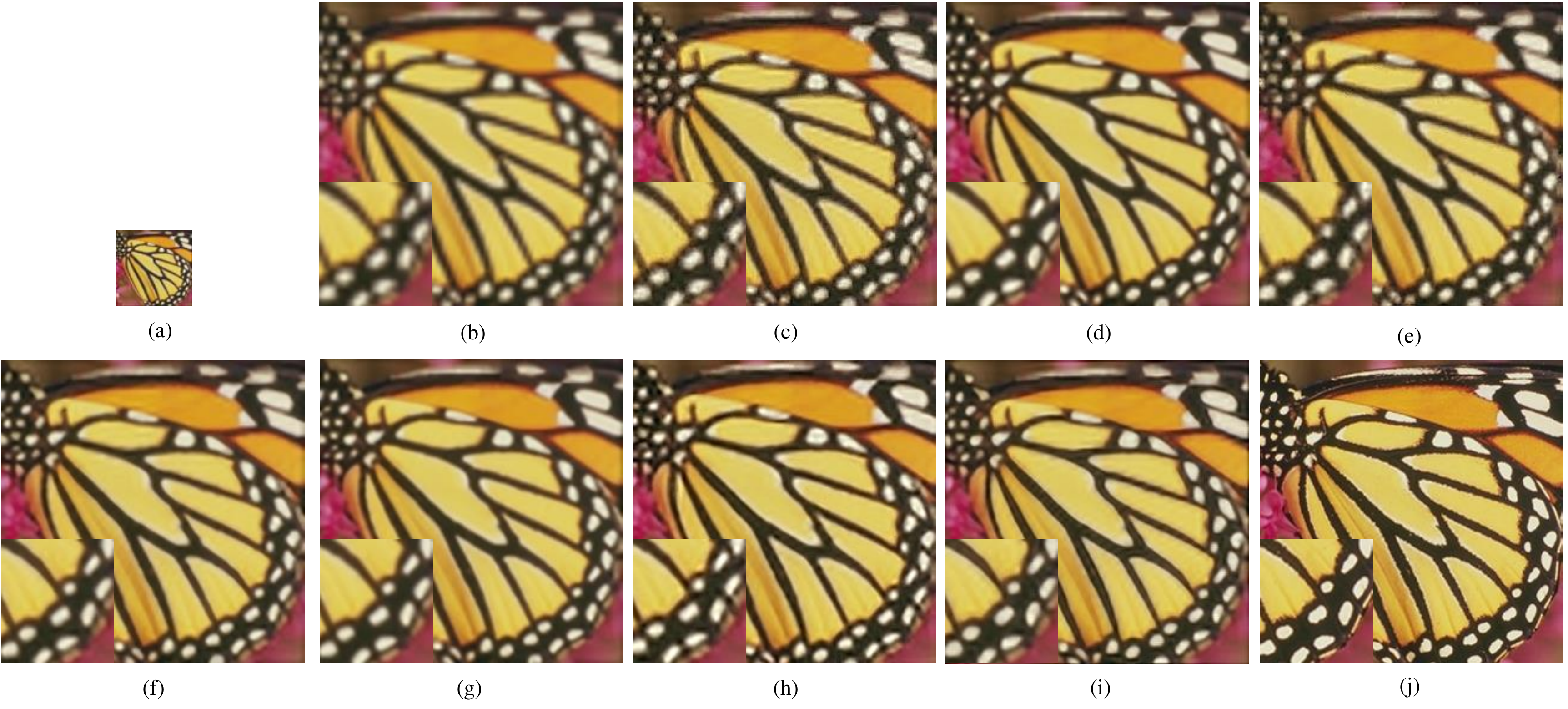}
\caption{Visual results comparison for image \emph{butterfly} ($\times4$, $\sigma=1.6$). (a)LR input. (b)bicubic. (c)SC-SR\cite{Yang2010}. (d)ASDS\cite{Dong2011}. (e)LRNE-SR\cite{Chen2014}. (f)DM-SR\cite{Bevilacqua2014}. (g)NCSR\cite{Dong2013}. (h)Aplus\cite{Aplus2015}. (i)proposed method. (j)ground truth.}
\label{fig:times4b}
\end{figure*}

\begin{figure*}[!htb]
\setlength{\abovecaptionskip}{-0.5cm}
\includegraphics[width=\textwidth]{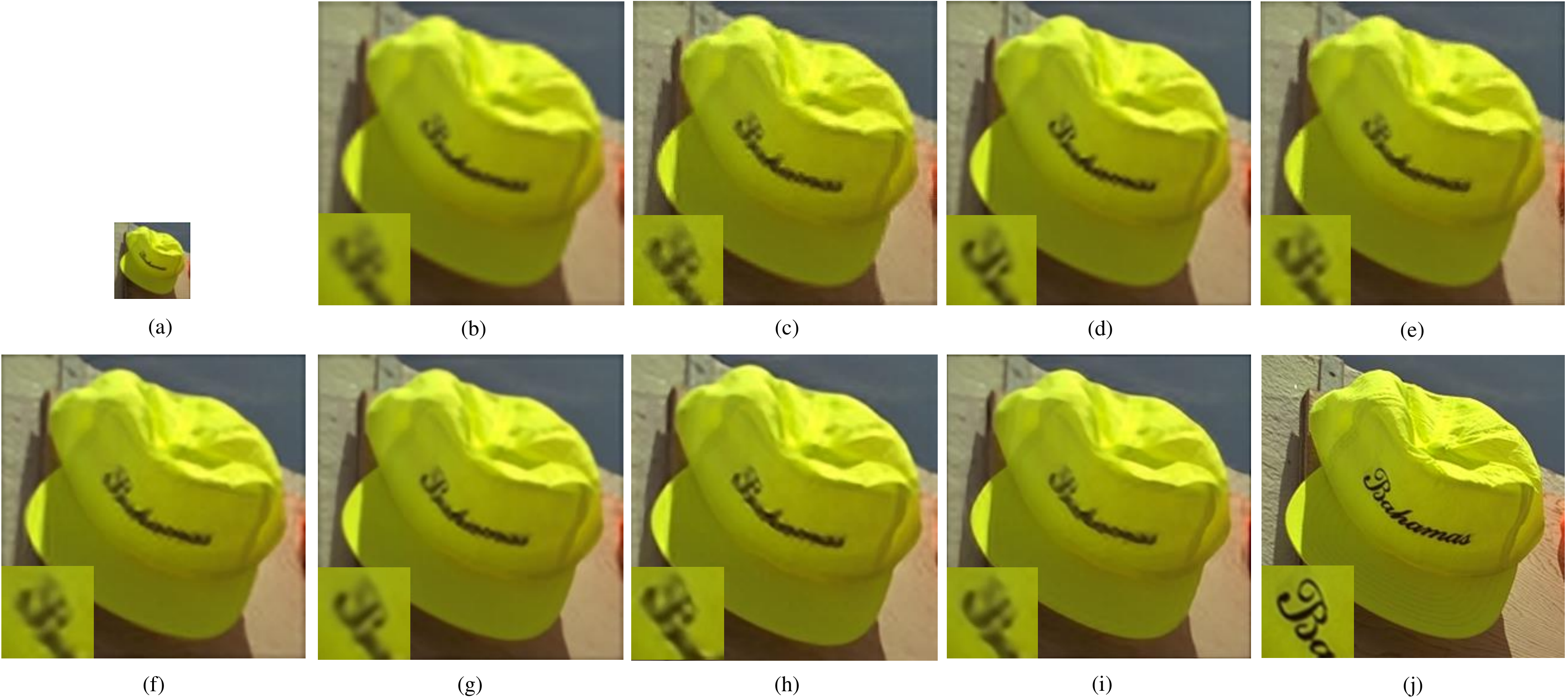}
\caption{Visual results comparison for image \emph{hat} ($\times4$, $\sigma=1.6$). (a)LR input. (b)bicubic. (c)SC-SR\cite{Yang2010}. (d)ASDS\cite{Dong2011}. (e)LRNE-SR\cite{Chen2014}. (f)DM-SR\cite{Bevilacqua2014}. (g)NCSR\cite{Dong2013}. (h)Aplus\cite{Aplus2015}. (i)proposed method. (j)ground truth.}
\label{fig:times4h}
\end{figure*}

 Fig. \ref{fig:times3g}-\ref{fig:times3h} show the visual SR results in the case of $p=3$ on the images \emph{girl} \emph{butterfly} and \emph{hat} by different methods, respectively. Fig. \ref{fig:times4g}-\ref{fig:times4h} show the visual SR results of the same test images with the factor of $p=4$. The SC-SR method \cite{Yang2010} generates blurry along edges (i.e., the boundary of the girl's nose) because the single over-complete dictionary learned from the external training images is not prone to produce sharp edges. LRNE-SR \cite{Chen2014} tends to lose high-frequency details while smooth regions and clean edges are produced. As one of the state-of-the-art methods for image SR, Aplus \cite{Aplus2015} obtains the second best quantitative SR performances (see Table \ref{tab:Table3} and \ref{tab:Table4}), however, too sharp boundaries and ringing artifacts can also be observed. Our proposed method generates obvious boundaries and suppresses artifacts. We can see clear edges of girl¡¯s nose and natural patterns in the wing of butterfly. As seen from the visual experimental results, our proposed method gets better results than other methods perceptually.

\subsection{Evaluation of the different contributions}
\par
\begin{figure}[!ht]
\setlength{\abovecaptionskip}{-0.6cm}
\includegraphics[width=\textwidth]{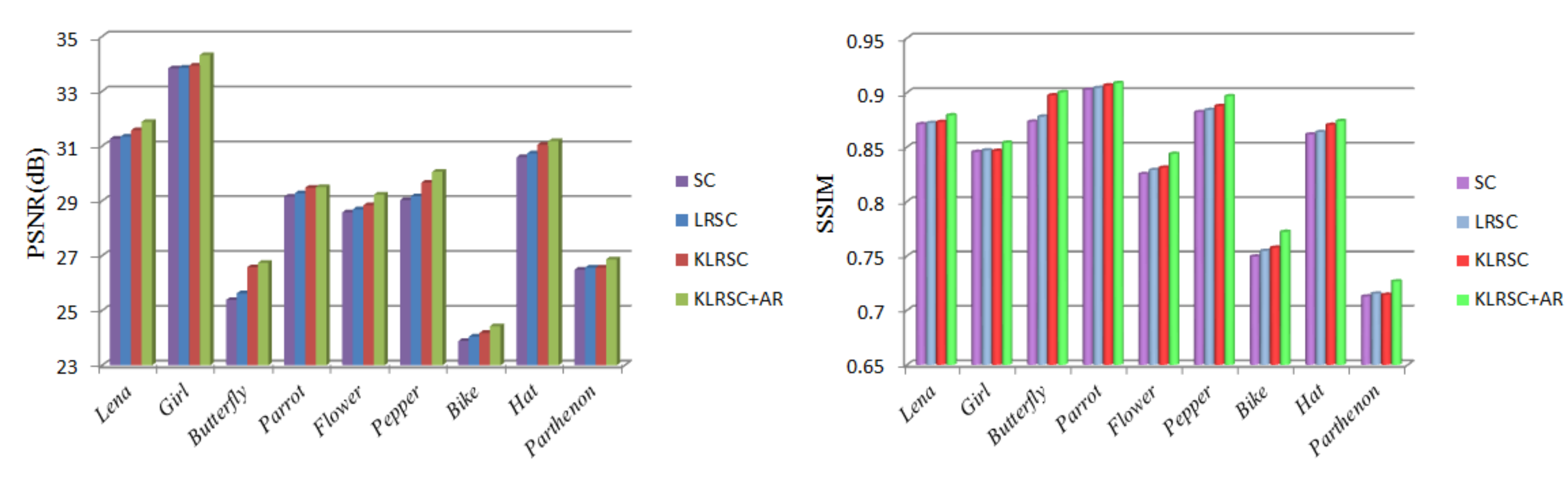}
\caption{SR performances  with the scaling factor 3 using sparse coding with different constraints and regularization.}
\label{fig:gain}
\end{figure}

To further validate the effectiveness of proposed method, we test the SR performances with the scaling factor $p=3$ using sparse coding with different constraints and regularization. The results are shown in Fig. \ref{fig:gain}. For the convenience of description, we denote sparse coding as 'SC', sparse coding with low-rank constraint as 'LRSC' and autoregressive as 'AR'. Our proposed KLRSC incorporated with AR model gains the highest PSNRs and SSIMs for all the reconstructions of images. Table \ref{tab:gain} shows the average SR performance using SC, LRSC, KLRSC and KLRSC+AR. The proposed KLRSC+AR method has an average improvement of 0.664dB in PSNR and 0.0148 in SSIM over the method using standard sparse representation, where the average PSNR/SSIM contributions of the low-rank constraint, the kernel method and the AR regularization are 0.120dB/0.0028, 0.284dB/0.0041 and 0.260dB/0.0079, respectively. It indicates that the incorporation of low-rank constraint, kernel method and AR regularization indeed boosts the SR results.
\begin{table}[!ht]
  \centering
  \caption{The average SR performance using SC, LRSC, KLRSC and KLRSC+AR.}
   \footnotesize
    \begin{tabular}{ccccccccc}
\toprule
    & \multicolumn{2}{c}{\multirow{2}[2]{*}{SC}}& \multicolumn{2}{c}{\multirow{2}[2]{*}{LRSC}} &\multicolumn{2}{c}{\multirow{2}[2]{*}{KLRSC}} & \multicolumn{2}{c}{proposed}\\
    &&&&&&&\multicolumn{2}{c}{KLRSC+AR} \\\cline{2-9}
    &PSNR&SSIM&PSNR&SSIM&PSNR&SSIM&PSNR&SSIM \\
    \toprule
   Avg. &28.692& 0.8358& 28.812 & 0.8386 & 29.096 & 0.8427 & \textbf{29.356} & \textbf{0.8506} \\
    \bottomrule
    \end{tabular}%
  \label{tab:gain}%
\end{table}%

\section{Conclusion}
 In this paper, we propose a novel single image SR method by incorporating self-similarity learning framework with kernel based low-rank sparse coding. The kernel method is used which captures the nonlinear structures of the input data. A novel kernel based low-rank sparse coding based scheme for single image SR is proposed, which exploits both the structural information of nonlocal-similarity in kernel space. Furthermore, we exploit the self-similarity redundancy among patches across different scales in a single natural image to train a self-example dictionary. The gradual magnification framework compatible to the self-example dictionary is adopted. Experimental results demonstrate that our proposed method improves SR performances both quantitatively and perceptually.

\section*{References}

\bibliography{mybibfile}

\end{document}